# Utilizing Domain Knowledge: Robust Machine Learning for Building Energy Prediction with Small, Inconsistent Datasets

Xia Chen, Manav Mahan Singh and Philipp Geyer

**Abstract**— The demand for a huge amount of data for machine learning (ML) applications is currently a bottleneck in an empirically dominated field. We propose a method to combine prior knowledge with data-driven methods to significantly reduce their data dependency. In this study, component-based machine learning (CBML) as the knowledge-encoded data-driven method is examined in the context of energy-efficient building engineering. It encodes the abstraction of building structural knowledge as semantic information in the model organization. We design a case experiment to understand the efficacy of knowledge-encoded ML in sparse data input (1% - 0.0125% sampling rate). The result reveals its three advanced features compared with pure ML methods: 1. Significant improvement in the robustness of ML to extremely small-size and inconsistent datasets; 2. Efficient data utilization from different entities' record collections; 3. Characteristics of accepting incomplete data with high interpretability and reduced training time. All these features provide a promising path to alleviating the deployment bottleneck of data-intensive methods and contribute to efficient real-world data usage. Moreover, four necessary prerequisites are summarized in this study that ensures the target scenario benefits by combining prior knowledge and ML generalization.

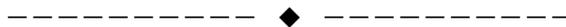

## 1 INTRODUCTION

In a review of the historical path of machine learning (ML) and artificial intelligence (AI), symbolism and connectionism have had many encounters [1], [2]. Besides traditional knowledge-based, symbolistic methods, the rapid development of AI in the past decade encourages many engineering tasks to use connectionist paradigms for prediction and simulation as decision-making support [3], [4], [5]. These models own good generalizations provided in those data-rich and pattern implicit domains, such as image classification [6], object detection [7], natural language processing [8], and even their combination: multimodality [9]. Besides their decent representation learning ability, AI-driven models achieved reliable results, even better than human performance, primarily supported by large datasets [10], [11], [12].

This prerequisite limits the application of current AI or data-driven methods into domains that comply with complex data-acquisition scenarios or empiricism contexts. These empiricism dominant domains are built on the foundation of highly integrated explicit laws and prior knowledge. In these domains, symbolism- or semantic-based methods are more well-acknowledged, such as the first-principles modeling via domain knowledge. As a highly compressed informative abstraction from human experience through observation and deduction (e.g., physics), knowledge exists with great extrapolation potential in empiricism dominant domains [13]. Therefore, embedding prior knowledge into ML is a direction worth exploring in depth to address the abovementioned issues.

The objective to combine data-driven models with knowledge-based methods in real-world engineering schemes starts from a practical perspective: fitting the scenario with a better generalization, robust accuracy, and effortless deployment. In the context of energy-efficient building design in the early stages, building performance simulation (BPS) and energy system dimensioning imply and validate a creational process of making geometry which causes a direct and fundamental impact on building performance [14], [15]. As an assessment tool, both ML and first-principles simulation are widely accepted by the community [16], [17], [18] but also come with their inherited limitations: Detailed simulation requires large engineering efforts and is time-consuming [19], [20], while the granularity of data-driven predictions is often positively correlated with the amount and quality of collected data [21], [22]. Engineers and designers need rapid and sufficiently accurate feedback from potential design space exploration (DSE) [23] to locate engineering solutions based on energy performance. Due to complex dependencies in energy-efficient sustainable design, the difficulty and cost of obtaining relevant data are in severe conditions or even lack appropriate data to support constructing sufficiently generalized models. Thus, easing the dependence of models on input data size is a critical factor in removing barriers to applying relevant research.

In this context, we hypothesize that combining engi-

---

- *Xia Chen and Philipp Geyer are with the Leibniz University Hannover, Institute for Design and Construction, Sustainable Building Systems Group, Hannover, 30419, Germany, E-mail: {xia.chen, philipp.geyer}@iek.uni-hannover.de.*
- *Manav Mahan Singh is with the Technical University of Munich, Georg Nemetschek Institute Artificial Intelligence for the Built World, Munich, 85748, Germany. E-mail: manavmahan.singh@tum.de.*

neering knowledge with data-driven models alleviates the performance limitation of data input requirements. We adopt a typical knowledge-encoded data-driven method in the building engineering domain: component-based machine learning (CBML) [24], [25] and design a series of performance evaluation experiments with different data sizes/sparseness in a BPS time-series prediction scenario. This study proves novelty two-fold:

1. Encoding semantic prior knowledge in the model organizational structure significantly contributes to:
   a. Robustness against accuracy decreases when input data are highly sparse, inconsistent, and size-wise small.
   b. Efficient data utilization from different entities' record collections.
   c. Characteristics of accepting incomplete data with high interpretability and reduced training time.
   d. It has substantial practical implications in reducing difficulties/costs in real-world industry deployment.
2. We extract four necessary prerequisites to ensure that combining prior engineering knowledge and ML generalization benefits the target scenario.

The rest of the content is organized as follows: Section 2 describes the research background with explanations of necessary methodologies in the scope of this study. The results with analysis are presented in Section 3. Section 4 discusses the adaptability, limitation, and future research outlook. Section 5 concludes the study.

## 2 BACKGROUND AND METHODOLOGIES

### 2.1 Engineering knowledge integration

Engineering inverse problems with hidden complex physics sometimes are effort-expensive for simulation or prediction [26], [27] because of grapy or noisy data conditions. It leads to the difficulties of solving "how" problems in semantic description through traditional approaches. ML has emerged as a promising alternative for solving these complex high-dimensional mesh fitting tasks. Studies point out that models trained from additional physical constraints or prior knowledge embedding have been introduced as physical-informed machine learning [26]. For the current data-driven process, expediency, transparency, and explainability requests have raised concerns and argue for rebalancing through a task-dependent symbiosis of fitting and interpreting [28].

A similar situation exists in the building engineering domain: inconsistent and missing data are prevalent in real-world scenarios [21], [22]. In this context, merging task-dependent assumptions in the cases through custom interventions in ML models is a way to embody physically relevant relationships for data augmentation. These assumptions satisfy a given set of physical laws or domain knowledge to ensure that the predictions sought reliable DSE [29]. By integrating prior knowledge into the data-driven process, we observed the improvement in training efficiencies [30], avoiding spurious causal conclusions [31], and extrapolation/generalization tasks improvement [32], [33].

In this study, our primary focus is integrating the extracted building semantic knowledge into the ML model organization. We set to investigate whether it enables reducing ML model dependency on data input, and whether it enhances the model robustness against data sparseness.

### 2.2 Component-based Machine Learning and Monolithic Architecture

A universal but knowledge-agnostic way to construct ML models to imitate BPS is by following the traditional monolithic architecture: a single unified model (internal layers and configurations could be varied) approximating the input-output relationship by learning implicit patterns in a black-box manner, as presented in Figure 1 (b). However, the problem that commonly arises in adapting ML prediction in BPS is that building variations are difficult to represent by a fixed set of numerical input parameters; It is impossible to represent all variations as numbers. Consequently, using a parametric-only approach, the model learns patterns only within a fixed building structure pattern, making it less transferable to new building shapes or geometries, hence, largely restricting its generalization. Essentially, it is less flexible for monolithic ML to acquire knowledge regarding the building topology in compact data input and map to its performance (output).

To overcome the above-mentioned challenge of building representation, embedding prior knowledge on the foundations of engineering-based decomposition of buildings is one of the solutions. Accordingly, we have developed a component-based machine learning that allows structural variations beyond parameterization to overcome the representation challenges of new buildings [25]. It follows the engineering-based decomposition of the building energy models into smaller, less physical complex, and more manageable components, such as walls, floors, zones, etc., constituting a set of generalized ML models to predict their individual performances (sub-figure c, Figure 1), composed as required to indicate the building energy performance [34]. In other words, this approach trains a transferable set of data-driven building component models that are applied to structurally different BPSs without retraining them for specific cases. From the perspective of problem space, the engineering prior knowledge decomposition gives a precious ability: it generalizes certain types of building design space into a broader set of building design variations space. This process converts extrapolation problems, which MLs are bad at [35], [36], into interpolation problems, with fewer data required. Figure 2 presents this conceptual illustration.

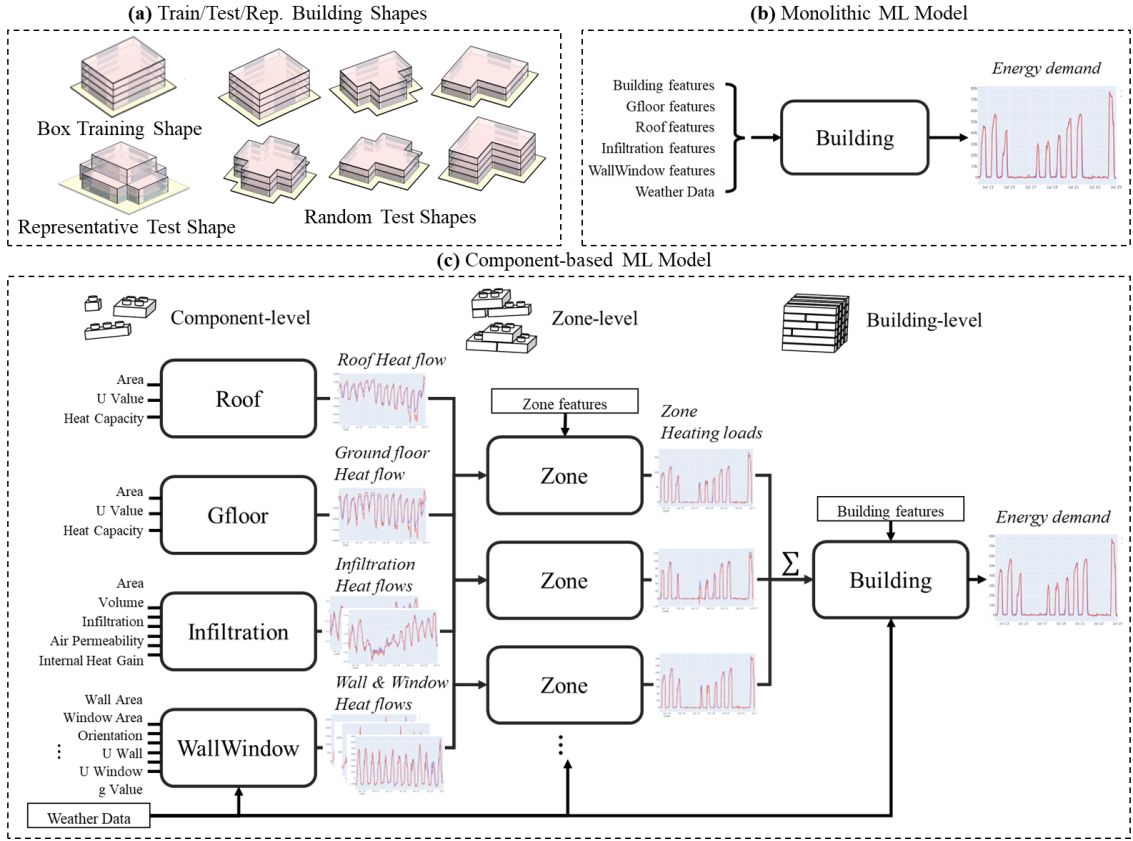

Fig. 1. Dataset illustration and two predictive modeling strategies (a) the dataset generated for training/test sets in different building shapes. In this study, we use a box-shape buildings dataset for model training using (b) monolithic ML model and (c) component-based ML model strategies. Both approaches are then validated on test datasets in box-shape, random shapes, and representative buildings.

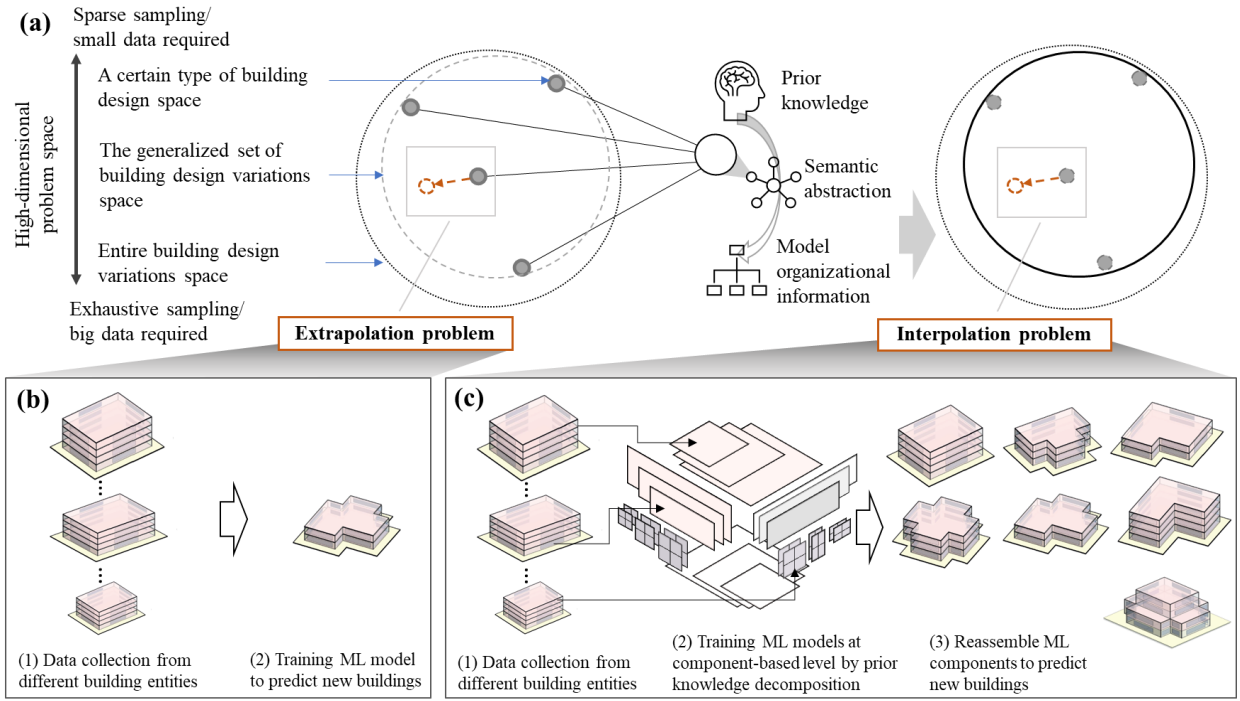

Fig. 2. Generalization via prior knowledge encoding: (a) The scope of problem space: The knowledge-encoded ML (CBML) allows data collection from different building instances (certain types of building design space), trains ML component models through building semantic abstraction, and predict new buildings by reassembling components; It transfers the ML prediction problem from extrapolation to interpolation and cover a larger problem space; (b) Monolithic modeling: extrapolation problem; (c) Component-based machine learning: interpolation problem; It overcomes the limitation of requiring complete input collection from buildings and maximizes the data utility.

Essentially, CBML inherits the generalization characteristics of ML and decouples the structured information through prior knowledge in the building engineering domain. Comparisons with the monolithic approach regarding its interpretability, transferability, and integration into the existing data structures have been discussed extensively [25], [37], [38]. We note that the effective knowledge extraction process fulfills the following two conditions:

- It involves structured semantic information and is representable through engineering prior knowledge which is universally applicable. In the scope of this study, semantic information originates from building information modeling [39]. The use of building performance simulation [40] adds engineering knowledge.
- The extracted knowledge is generally applicable to different buildings by the use of components recurring in the majority of designs.

Since the knowledge-encoded process extends the problem space with the same data samples, correspondingly, it raises the sparseness of the data. Hence, the problem is necessary to investigate how robust the CBML organization deals with the sparse data.

### 2.3 Data Description

To conduct a case experiment, we investigate the building performance data in time-series format instead of point prediction for two reasons:

1. Time-series data carries more detailed information regarding building performance peak load, behavior patterns, and dynamic factor analysis. It is essential in supporting building engineering optimization in both design and operation phases.
2. Time-series format ensures sets of sub-data well cover each building sample. It avoids too dense sampling within building parameters range and potential overfitting in ML training.

We adapted a validated dataset widely investigated from previous research [33], [41], [42]: Different building representation levels (components, zone, and building) in geometrically different building shapes sampled in set parameter ranges. The dataset is categorized into three subsets based on the geometry difference: *Box*, *Random*, and *Representative* (subfigure a, Figure 1). We extended from point estimate to hourly time-series performance by setting the target outputs *Heating Load* in a *typical winter period* between January 05 – January 17 in Munich, Germany. Accordingly, a semantic input/output illustration of monolithic ML and CBML adapted in time-series prediction and three building shape categories illustrations follows the presentation of Figure 1. We generated rich time-series data in 5000 box-shape building samples for training, and 1000 samples for box, random and representative shapes for the test (Table 1). The detailed dataset description, data generation process, and parameter ranges are available in Appendix.

### 2.4 ML Algorithm: LightGBM

The building parametric representation is a complex, high-dimensional multivariate encoding hyperspace. The modern ML exposes the power of well interpolating performance in approximation over this hyperspace. In fact, for recognizing implicit dynamics in time-series data in the state-of-the-art, reviews [43], [44] point out two current prominent advanced algorithm families applied in our domain: neural networks (NNs) and boosting algorithms. The reason behind it is two-fold: On the one hand, both approaches own higher capacities by progressively deconstructing the problem and fitting it into vast parametric representation (tree nodes split and neurons weight) instead of determining complex hyperplane to separate data (e.g. support vector machine or logistic regression); On the other hand, training to minimize loss by gradient descent and regularizations are widely adapted into NNs and boosting algorithms, makes them easy to transfer various tasks into optimization problems for convergence. We summarized two necessary features as ML algorithm selection:

- Generalization capabilities within a high-dimensional encoding space problem.
- Effective objective function(s) and regularization for efficient convergence.

To ensure optimal performance, NNs usually require case-based network structure design, fine-tuning in features of neural nodes, layers, and activation functions according to the data structure, size, etc. In comparison, boosting algorithms, especially for time-series prediction, require less extensive preprocessing or tuning to perform accurately with generalization flexibility [43], [45]. In this study, we select Light Gradient Boosting Machine (LightGBM) as our ML model. Further insight and open-source algorithm implementation in detail are available in the original paper [46].

### 2.5 Experiment Design and Training Strategy

TABLE 1
DATA SIZES OF DIFFERENT SETS. THE ORIGINAL TRAIN/TEST DATASET CONTAINS 5000/1000 VARIATE BUILDING SAMPLES, EACH BUILDING OWNS CORRESPONDING COMPONENTS WITH 312 HOURLY DATA POINTS FROM THE TARGET PERIOD.

|  | Number of data size | | | |
| --- | --- | --- | --- | --- |
| **ML Component** | **Box/Train** | **Box/Test** | **Random** | **Representation** |
| *Building* | 1,560,000 | 312,000 | 312,000 | 312,000 |
| *Zone / Infiltration* | 5,460,000 | 1,092,312 | 1,092,000 | 1,248,000 |
| *Ground Floor* | 1,560,000 | 312,000 | 312,000 | 312,000 |
| *Roof* | 1,560,000 | 312,000 | 312,000 | 1,560,000 |
| *Wall and Window* | 21,840,000 | 4,369,248 | 6,746,688 | 10,608,000 |

The objective of the experiment design is to inspect two major characteristics of the knowledge-encoded ML model (CBML): generalization ability and robustness against data sparseness. The evaluation of model generalization capacities is designed in the model train-test process, we trained both CBML component ML models and the monolithic building ML model on the box-shape training set mentioned above. The performance of trained models is then further validated in three building-shape test sets: box, random, and representative. The same process is conducted with different reduced training data sized via random sampling. To simulate sparse, inconsistent data, we sampled randomly in the training set with a high masking ratio of 1%, 0.1%, 0.05%, 0.025%, and 0.0125%, individually, thus creating a task that cannot be easily solved by a pure data-driven ML approach. At the building level, the smallest training set only contains 195 inconsistent time-series hourly data points out of 5000 different building samples. The validation process remains the same with full-scale test sets.

The training strategy of ML models aims to avoid biased results caused by model over-/underfitting. The setting of most hyperparameters[1] are kept as default but we fine-tune the model iteration round. To utilize the full-scale dataset and model capability, we used 3-fold cross-validation in the training process with 300 early stopping epochs. The averaged error was under supervised during the cross-validation training process until it stops dropping for 300 iterations, then the best iteration rounds are confirmed.

## 3 RESULTS

Results showing the performance of CBML and monolithic ML with different data sparseness are presented in Figure 3 and Table 2. As the prior knowledge is embedded into the ML model, we observed serval advantageous features based on result evaluation.

**Robustness against data sparseness:** With an increasing design complexity of the test set (shape of box, random, and representative), the performances of both CBML and monolithic modeling approaches decrease slightly, accordingly. This trend is intensified when the data sparseness grows. When the data input is rich (above 1% sampling rate), accuracies of both approaches are above 0.85 ($R^2$), with the monolithic ML model advancing slightly due to its more compact data input/output structure. The situation changes when the sampling rate drops to 0.1%. With more sparse training data, the performance difference between the monolithic ML model and the CBML model in all three test sets widens. Eventually, the accuracy of CBML models remains 0.6-0.7, while the accuracy of the monolithic model drops to 0.35-0.45. Although such low level is, of course, unacceptable for a prediction model, the observable trend with the increasing sparseness of data and the robustness of CBML is essential.

**Model interpretability: Table 2** gives more insight into the performances of ML components. We noticed that the increasing data sparseness does not affect ground floor, infiltration, and roof ML components, much. Their accuracy is above 0.9 at a 0.05% sampling rate and remains around 0.7 at a sampling rate of 0.0125%. In comparison, the wall & window component is more sensitive to the data sparseness and the building's design complexity as the thermal behavior of the component is more complicated. It is more directly affected by external conditions, orientation, building geometry, etc. Under the sampling rate of 0.0125%, we observe that the ML model completely fails to learn the pattern (-0.04 of $R^2$); Even for higher sampling rates, the accuracy is critical. The decoupling of building semantic information from data allows building designers and engineers to access information at the component level and make reasoning for a better informative support.

**Robustness against noisy/incomplete input:** Additionally, even with the noisy input of wall & window components (negative $R^2$ in the last column of Table 2), the accuracy of the CBML model is still maintained at 0.63. Compared to 0.39 accuracy in the monolithic ML model, CBML certainly gives a more robust characteristic not only against data sparseness but also against noisy information from the component level. Since the wall and window components at the test set of the representative building give almost pure noise at a 0.0125% sampling rate, it means that CBML remains robust when a certain level of incomplete inputs is replaced with random input.

**Training efforts:** Finally, it's worth emphasizing that less data requirement also means less time and effort investment in training. The organization CBML alleviates the difficulty of the training/tuning process from a compact monolithic model to adjusting by components, separately. Interchangeable characteristics of components enable each to be utilized, operated, and maintained by small datasets. This characteristic relieves the effort of constructing a general model that involves data from different building cases. Naturally, a reduced data size corresponds to a shortening of the training process time.

To sum up, in sparse data input, the abstraction of building domain knowledge and encoding it as model organizational information improves the performance robustness of ML models significantly. It further endows the model with characteristics of accepting incomplete data with high interpretability and less training time. These features play a pivotal beneficial role in alleviating deployment difficulties in solving actual engineering tasks.

## 4 DISCUSSION

Our experiment results prove that the knowledge-encoded ML model organization, component-based machine learning, is more robust against data sparseness and noisy data input. At the same time, it shows better model interpretability and shorter training time. More im-portantly, its organization accepts sparse data collections from different building entities and integrates them as a whole for prediction. Thus, it offers an intuitive modeling framework that scales well with fewer data requirements which is the core of addressing engineering modeling tasks in deployment. Within the scope of the building engineering domain, this study shows how a component-based machine learning approach extracts the building domain knowledge and transfers it into the structure of the ML model. We note that as a medium of carrying implicit patterns, data owns the potential to be decoupled via domain knowledge and carries less complex information to mitigate the learning costs of ML models.

---
[1] Default setting of hyperparameters is available on LightGBM 3.2.1.99 documentation (2021):
https://lightgbm.readthedocs.io/en/latest/Parameters.html?highlight=default

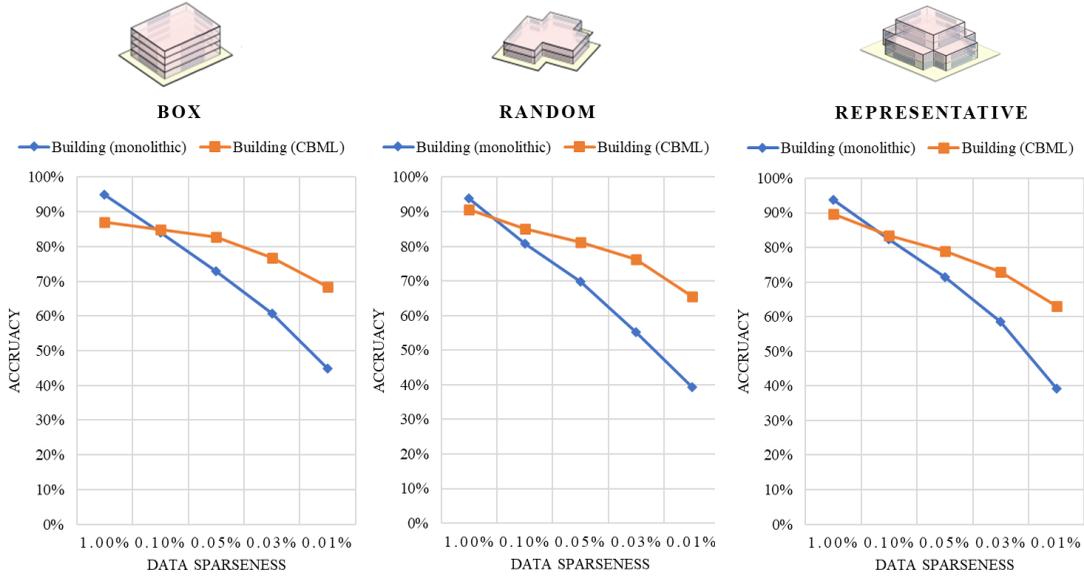

Fig. 3. Accuracy result comparison on different test datasets. Models are trained on different sparseness of the training data (sampling rate) from 1.00% to 0.0125%.

TABLE 2
PREDICTION ACCURACY ($R^2$) RESULTS OF ML MODELS VALIDATED ON DIFFERENT TEST DATA WITH VARIATE TRAINING DATA SPARSENESS.

| Training data sparseness/size | 1.00%/15,600 | | | 0.10%/1,560 | | | 0.05%/780 | | | 0.025%/390 | | | 0.0125%/195 | | |
|---|---|---|---|---|---|---|---|---|---|---|---|---|---|---|---|
| Test data | Box | Rand. | Rep. | Box | Rand. | Rep. | Box | Rand. | Rep. | Box | Rand. | Rep. | Box | Rand. | Rep. |
| *Building\** | **0.95** | **0.94** | **0.94** | 0.84 | 0.81 | 0.82 | 0.73 | 0.70 | 0.71 | 0.61 | 0.55 | 0.59 | 0.45 | 0.39 | 0.39 |
| *CBML Building* | 0.88 | 0.91 | 0.90 | **0.85** | **0.85** | **0.83** | **0.83** | **0.81** | **0.79** | **0.77** | **0.76** | **0.73** | **0.68** | **0.66** | **0.63** |
| *Ground floor* | 0.94 | 0.93 | 0.94 | 0.92 | 0.90 | 0.89 | 0.89 | 0.87 | 0.87 | 0.83 | 0.80 | 0.80 | 0.70 | 0.68 | 0.70 |
| *Infiltration* | 0.96 | 0.97 | 0.95 | 0.93 | 0.93 | 0.94 | 0.92 | 0.92 | 0.89 | 0.89 | 0.88 | 0.83 | 0.85 | 0.85 | 0.81 |
| *Roof* | 0.95 | 0.96 | 0.96 | 0.94 | 0.94 | 0.94 | 0.91 | 0.93 | 0.92 | 0.88 | 0.87 | 0.87 | 0.70 | 0.69 | 0.69 |
| *Wall & Window* | 0.98 | 0.95 | 0.82 | 0.91 | 0.85 | 0.66 | 0.84 | 0.73 | *0.39* | 0.74 | 0.59 | *0.12* | 0.63 | 0.46 | *-0.04* |

\* *using monolithic ML*

Essentially, the knowledge extraction from the data is a rebalancing process between semantic domain information and input noise: The knowledge-encoding process brings organization information, but it might ignore some hidden patterns by process simplification and introduce uncertainties in the framework. Because of its multi-component structure organization, uncertainties and noise might be accumulated through different levels of components in CBML and lead to limited accuracy [42]. The monolithic model captures the implicit patterns well when the data condition is rich and the prediction task is within a dense data range. However, with the increasing data sparseness, we observe that monolithic ML starts to fail to learn the pattern from insufficient data. In this data-sparse, inconsistent situation, the knowledge encoded in model organizational structure (CBML) brings the advantage in accuracy robustness. It is reasonable to foresee that more efforts in domain knowledge extraction and encoding technics developments would contribute to compensating for the data-hungry deficiency of ML and using available data more efficiently.

Additionally, we note that the break-even point that CBML outperforms monolithic ML is when the resampled data size is less than the building samples (5000 building samples in the training set). It means the model doesn't receive certain building geometry types and their performance patterns at all. The result validates what subfigure (a) in Figure 1 described: The semantic information extracted from domain knowledge helps CBML go beyond application context in the training data without extrapolating at the component level. Based on this finding, we can imagine another different combination: By recognizing the time-series patterns of building thermo-dynamic behavior as domain knowledge, it is possible to transfer point building performance prediction to time-series forecasting by combing building performance sequential patterns, knowledge, and the point estimation ML model. As we see, utilizing the generalization ability of the ML model and the domain knowledge abstraction could generate various benefits for task-solving assistance for designers and engineers. This flexible characteristic certainly contributes to adopting this framework to fit broader domains and scenarios.

Finally, by recognizing the fit of the knowledge-encoded data-driven method, we summarize four prerequisites based on this study to ensure the target scenario would benefit from combining domain knowledge and ML generalization:

1. The task involves structured information that is representable by domain knowledge.
2. The task data is widely but not necessarily dense sampled.
3. The model owns the capacity for exploring the task's high-dimensional hyperspace.
4. The model fits to transfer task to correct objective function and effective regularization to converge.

Primaries (1) and (2) ensure that the domain knowledge is extractable and representable. (3) and (4) set the features the ML model should fulfill.

## 7 CONCLUSIONS

In this study, we introduce the rebalancing of integrating prior knowledge into the data-driven processes in the building engineering domain. The designed experiment proves that the way of encoding semantic information into component-based machine learning overcomes the difficulties of data reliance in single monolithic ML models. In the context of building engineering tasks, such a knowledge-encoded mindset releases the possibility to adopt data from different buildings and serve as an informed decision-support assistant in new buildings. More importantly, our study proves that even trained in sparse, inconsistent, and small-size datasets, the knowledge-encoded, component-based machine learning presents a more robust prediction accuracy. We consider this finding as a significant impact on alleviating data-hungry difficulties in practical data-driven model deployment and contributes to efficient real-world data usage in engineering transferable to all fields that match the mentioned prerequisites.

## ACKNOWLEDGMENTS

We acknowledge the German Research Foundation (DFG) support for funding the project under grant GE 1652/3-2 in the Researcher Unit FOR 2363 and under Heisenberg grant GE 1652/4-1.

## REFERENCES

[1] M. L. Minsky, "Logical versus analogical or symbolic versus connectionist or neat versus scruffy," AI magazine, vol. 12, no. 2, p. 34, 1991.

[2] M. Garnelo and M. Shanahan, "Reconciling deep learning with symbolic artificial intelligence: representing objects and relations," Current Opinion in Behavioral Sciences, vol. 29, pp. 17–23, 2019.

[3] N. D. Roman, F. Bre, V. D. Fachinotti, and R. Lamberts, "Application and characterization of metamodels based on artificial neural networks for building performance simulation: A systematic review," Energy and Buildings, vol. 217, p. 109972, 2020, doi: 10.1016/j.enbuild.2020.109972.

[4] C. Zhang and Y. Lu, "Study on artificial intelligence: The state of the art and future prospects," Journal of Industrial Information Integration, vol. 23, p. 100224, 2021.

[5] Y. Duan, J. S. Edwards, and Y. K. Dwivedi, "Artificial intelligence for decision making in the era of Big Data - evolution, challenges and research agenda," International journal of information management, vol. 48, pp. 63–71, 2019.

[6] W. Rawat and Z. Wang, "Deep convolutional neural networks for image classification: A comprehensive review," Neural computation, vol. 29, no. 9, pp. 2352–2449, 2017.

[7] Z.-Q. Zhao, P. Zheng, S. Xu, and X. Wu, "Object detection with deep learning: A review," IEEE transactions on neural networks and learning systems, vol. 30, no. 11, pp. 3212–3232, 2019.

[8] D. Khurana, A. Koli, K. Khatter, and S. Singh, "Natural language processing: State of the art, current trends and challenges," Multimedia Tools and Applications, pp. 1–32, 2022.

[9] T. Baltrušaitis, C. Ahuja, and L.-P. Morency, "Multimodal machine learning: A survey and taxonomy," IEEE transactions on pattern analysis and machine intelligence, vol. 41, no. 2, pp. 423–443, 2018.

[10] A. Paullada, I. D. Raji, E. M. Bender, E. Denton, and A. Hanna, "Data and its (dis) contents: A survey of dataset development and use in machine learning research," Patterns, vol. 2, no. 11, p. 100336, 2021.

[11] M. Weber, M. Engert, N. Schaffer, J. Weking, and H. Krcmar, "Organizational Capabilities for AI Implementation—Coping with Inscrutability and Data Dependency in AI," Information Systems Frontiers, pp. 1–21, 2022.

[12] A. L'heureux, K. Grolinger, H. F. Elyamany, and M. am Capretz, "Machine learning with big data: Challenges and approaches," Ieee Access, vol. 5, pp. 7776–7797, 2017.

[13] P. Humphreys, Extending ourselves: Computational science, empiricism, and scientific method: Oxford University Press, 2004.

[14] T. L. Hemsath, "Conceptual energy modeling for architecture, planning and design: Impact of using building performance simulation in early design stages," in 13th Conference of International Building Performance Simulation Association, 2013, pp. 376–384.

[15] K. Negendahl, "Building performance simulation in the early design stage: An introduction to integrated dynamic models," Automation in Construction, vol. 54, pp. 39–53, 2015.

[16] P. Westermann and R. Evins, "Surrogate modelling for sustainable building design – A review," Energy and Buildings, vol. 198, pp. 170–186, 2019, doi: 10.1016/j.enbuild.2019.05.057.

[17] T. Han, Q. Huang, A. Zhang, and Q. Zhang, "Simulation-based decision support tools in the early design stages of a green building—A review," Sustainability, vol. 10, no. 10, p. 3696, 2018.

[18] T. Østergård, R. L. Jensen, and S. E. Maagaard, "Building simulations supporting decision making in early design – A review," Renewable and Sustainable Energy Reviews, vol. 61, pp. 187–201, 2016, doi: 10.1016/j.rser.2016.03.045.

[19] E. Azar et al., "Simulation-aided occupant-centric building design: A critical review of tools, methods, and applications," Energy and Buildings, vol. 224, p. 110292, 2020.

[20] A. Nutkiewicz, Z. Yang, and R. K. Jain, "Data-driven Urban Energy Simulation (DUE-S): A framework for integrating engineering simulation and machine learning methods in a multi-scale urban energy modeling workflow," Applied Energy, vol. 225, pp. 1176–1189, 2018, doi: 10.1016/j.apenergy.2018.05.023.

[21] W. Tian et al., "A review of uncertainty analysis in building energy assessment," Renewable and Sustainable Energy Reviews, vol. 93, pp. 285–301, 2018.

[22] B. R. K. Mantha, C. C. Menassa, and V. R. Kamat, "A taxonomy of data types and data collection methods for building energy monitoring and performance simulation," Advances in Building Energy Research, vol. 10, no. 2, pp. 263–293, 2016.

[23] T. Østergård, R. L. Jensen, and S. E. Maagaard, "Early Building Design: Informed decision-making by exploring multidimensional design space using sensitivity analysis," Energy and Buildings, vol. 142, pp. 8–22, 2017.

[24] P. Geyer, "Component-oriented decomposition for multidisciplinary design optimization in building design," Advanced Engineering Informatics, vol. 23, no. 1, pp. 12–31, 2009, doi: 10.1016/j.aei.2008.06.008.

[25] P. Geyer and S. Singaravel, "Component-based machine learning for performance prediction in building design," Applied Energy, vol. 228, pp. 1439–1453, 2018, doi: 10.1016/j.apenergy.2018.07.011.

[26] G. E. Karniadakis, I. G. Kevrekidis, L. Lu, P. Perdikaris, S. Wang, and L. Yang, "Physics-informed machine learning," Nature Reviews Physics, vol. 3, no. 6, pp. 422–440, 2021.

[27] A. G. Ramm, Inverse problems: mathematical and analytical techniques with applications to engineering: Springer Science & Business Media, 2006.

[28] J. Pearl, "Radical empiricism and machine learning research," Journal of Causal Inference, vol. 9, no. 1, pp. 78–82, 2021.

[29] J. Drgoňa, A. R. Tuor, V. Chandan, and D. L. Vrabie, "Physics-constrained deep learning of multi-zone building thermal dynamics," Energy and Buildings, vol. 243, p. 110992, 2021.

[30] X. Chen, T. Guo, M. Kriegel, and P. Geyer, "A hybrid-model forecasting framework for reducing the building energy performance gap," Advanced Engineering Informatics, vol. 52, p. 101627, 2022.

[31] X. Chen, J. Abualdenien, M. M. Singh, A. Borrmann, and P. Geyer, "Introducing causal inference in the energy-efficient building design process," arXiv preprint arXiv:2203.10115, 2022.

[32] S. Singaravel, J. Suykens, H. Janssen, and P. Geyer, "Explainable deep convolutional learning for intuitive model development by non‐machine learning domain experts," Design Science, vol. 6, 2020.

[33] P. Geyer, M. M. Singh, and X. Chen, "Explainable AI for engineering design: A unified approach of systems engineering and component-based deep learning," arXiv preprint arXiv:2108.13836, 2021.

[34] M. M. Singh, S. Singaravel, and P. Geyer, "Machine learning for early stage building energy prediction: Increment and enrichment," Applied Energy, vol. 304, p. 117787, 2021.

[35] A. Malistov and A. Trushin, "Gradient boosted trees with extrapolation," in 2019 18th IEEE International Conference On Machine Learning And Applications (ICMLA), 2019, pp. 783–789.

[36] P. J. Haley and D. Soloway, "Extrapolation limitations of multilayer feedforward neural networks," in [Proceedings 1992] IJCNN international joint conference on neural networks, 1992, pp. 25–30.

[37] P. Geyer, M. M. Singh, and S. Singaravel, "Component-Based Machine Learning for Energy Performance Prediction by MultiLOD Models in the Early Phases of Building Design," in Lecture Notes in Computer Science, Advanced Computing Strategies for Engineering, I. F. C. Smith and B. Domer, Eds., Cham: Springer International Publishing, 2018, pp. 516–534.

[38] M. M. Singh, S. Singaravel, R. Klein, and P. Geyer, "Quick energy prediction and comparison of options at the early design stage," Advanced Engineering Informatics, vol. 46, p. 101185, 2020.

[39] A. Borrmann, M. König, C. Koch, and J. Beetz, "Building information modeling: Why? what? how?," in Building information modeling: Springer, 2018, pp. 1–24.

[40] J. Hensen and R. Lamberts, "Building performance simulation for design and operation," 2019.

[41] M. M. Singh, C. Deb, and P. Geyer, "Early-stage design support combining machine learning and building information modelling," Automation in Construction, vol. 136, p. 104147, 2022.

[42] X. Chen, M. M. Singh, and P. Geyer, "Component-based machine learning for predicting representative time-series of energy performance in building design," in 28th International Workshop on Intelligent Computing in Engineering, Berlin, 2021.

[43] S. Papadopoulos, E. Azar, W.-L. Woon, and C. E. Kontokosta, "Evaluation of tree-based ensemble learning algorithms for building energy performance estimation," Journal of Building Performance Simulation, vol. 11, no. 3, pp. 322–332, 2018, doi: 10.1080/19401493.2017.1354919.

[44] D. Chakraborty and H. Elzarka, "Advanced machine learning techniques for building performance simulation: a comparative analysis," Journal of Building Performance Simulation, vol. 12, no. 2, pp. 193–207, 2019.

[45] S. Makridakis, F. Petropoulos, and E. Spiliotis, The M5 competition: Conclusions: Elsevier, International Journal of Forecasting.

[46] Guolin Ke et al., "LightGBM: A Highly Efficient Gradient Boosting Decision Tree,"

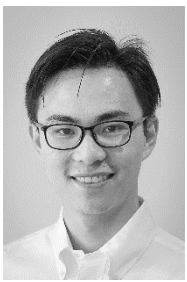

**Xia Chen** – PhD candidate at Leibniz University Hannover (Germany). His research field is sustainable building design, predictive modeling supported by digital simulation and artificial intelligence, integrating prior knowledge (first-principles methods) into data-driven approaches, causal inference analysis, and exploring new patterns for human-computer interaction.

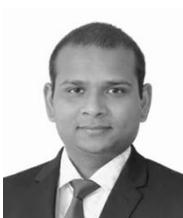

**Manav Mahan Singh** received his bachelor's degree in 2012, master's degree in 2014 and doctoral degree in 2022 from KU Leuven, Belgium. He has received GATE Scholarship (India), DAAD Scholarship (Germany), and PhD and ASL Scholarship (Belgium). He has won two national level awards - 3D Student Design Challenge, Autodesk, 2013, India and Digital Construction Brussels, 2018, Belgium. He has a rich experience of architecture and computer science from both industry and academia. His works at TUM Georg Nemetschek Institute, Germany, focusses on building information modelling and machine learning based design tools to support energy-efficient building design.

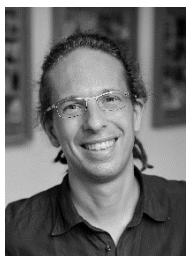

**Philipp Geyer** is Heisenberg Professor for Sustainable Building Systems at Leibniz University Hannover (Germany). His research field is sustainable building design, construction and technology supported by digital modelling, simulation, and intelligent computation, especially artificial intelligence and machine learning. He is chairing the European Group for Intelligent Computing in Engineering (eg-ice.org). Previously, he held positions at KU Leuven (Belgium), TU Munich (Germany), and ETH Zurich (Switzerland) and was visiting researcher at Massachusetts Institute of Technology (MIT). Furthermore, he has more than 80 peer-reviewed publications in major international journals, books, and conference proceedings.